# A Combination of Cutset Conditioning with Clique-Tree Propagation in the Pathfinder System

**H.J. Suermondt, G.F. Cooper, and D.E. Heckerman**
Medical Computer Science Group
MSOB X-215, Stanford University, Stanford, CA 94305-5479

*Abstract* - Cutset conditioning and clique-tree propagation are two popular methods for performing exact probabilistic inference in Bayesian belief networks. Cutset conditioning is based on decomposition of a subset of network nodes, whereas clique-tree propagation depends on aggregation of nodes. We describe a means to combine cutset conditioning and clique-tree propagation in an approach called *aggregation after decomposition (AD)*. We discuss the application of the AD method in the Pathfinder system, a medical expert system that offers assistance with diagnosis in hematopathology.

## 1. Introduction

The Pathfinder expert system resulted from a joint project by the University of Southern California and Stanford University to assist general pathologists with diagnosis in the specialty area of hematopathology [Heckerman, 1985; Heckerman, 1989; Heckerman, 1990b]. Uncertain knowledge must be represented and reasoned with in this system; therefore, the Pathfinder researchers have explored a variety of techniques for reasoning under uncertainty. After early failures to represent complex uncertain knowledge effectively with production rules and with nonprobabilistic scoring schemes, the Pathfinder team concentrated on decision-theoretic methods for diagnosis [Heckerman, 1990b]. Thus, the current representation of the Pathfinder knowledge base includes a multiply connected Bayesian belief network. In this belief network, 63 malignant and benign diseases of lymph nodes are represented by a single node, the *disease node*,[1] and over 100 morphologic and nonmorphologic *features* (findings) visible in lymph-node tissue are each represented by a separate node (see Figure 1). The feature nodes each have between two and ten mutually exclusive and exhaustive *values*. The disease node has 63 values, each representing a possible diagnosis. Thus, the underlying assumption in this representation is that diseases (but not features) in hematopathology are mutually exclusive, and that the set of possible diagnoses represented by the disease node is exhaustive. The assumption in Pathfinder that diseases in hematopathology are mutually exclusive is appropriate, because co-occurring diseases almost always appear in different lymph nodes or in different regions of the same lymph node, and a user can analyze each area of pathology separately.

The inference problem in the Pathfinder system is to calculate the marginal probability distribution for the disease node, given that various feature nodes have been instantiated to observed values. Exact probabilistic inference in Bayesian belief networks is NP-hard [Cooper, 1990]. Thus, it is quite unlikely that an algorithm for probabilistic inference on belief networks can ever be developed that is efficient even in the worst cases.

Recently, we chose to approach the Pathfinder inference problem with an algorithm developed by Lauritzen and Spiegelhalter [Lauritzen, 1988], later refined by the MUNIN team [Andersen, 1989]. We shall refer to this algorithm as *clique-tree propagation (CTP)*. The underlying principle of CTP is aggregation of nodes into clusters called *cliques*. Once the nodes have been grouped into cliques, the resulting structure is a tree, in which probabilities are propagated by local operations. Within the clique tree, we propagate evidence by first calculating the joint probability of the nodes in each clique, and then adjusting these joint probabilities to be consistent with those of adjacent cliques in the clique tree [Andersen, 1989]. The computational time complexity of CTP increases exponentially with the number of nodes in each clique [Lauritzen, 1988, pp. 186–188]: The complexity is proportional to the *state-space size* of each clique, which is the product of the numbers of possible values of the nodes in that clique. Thus, if inference is to be tractable, we must keep the cliques small.

---

[1] In this paper, we shall use the term *diagnosis* to indicate a single value of the disease node.

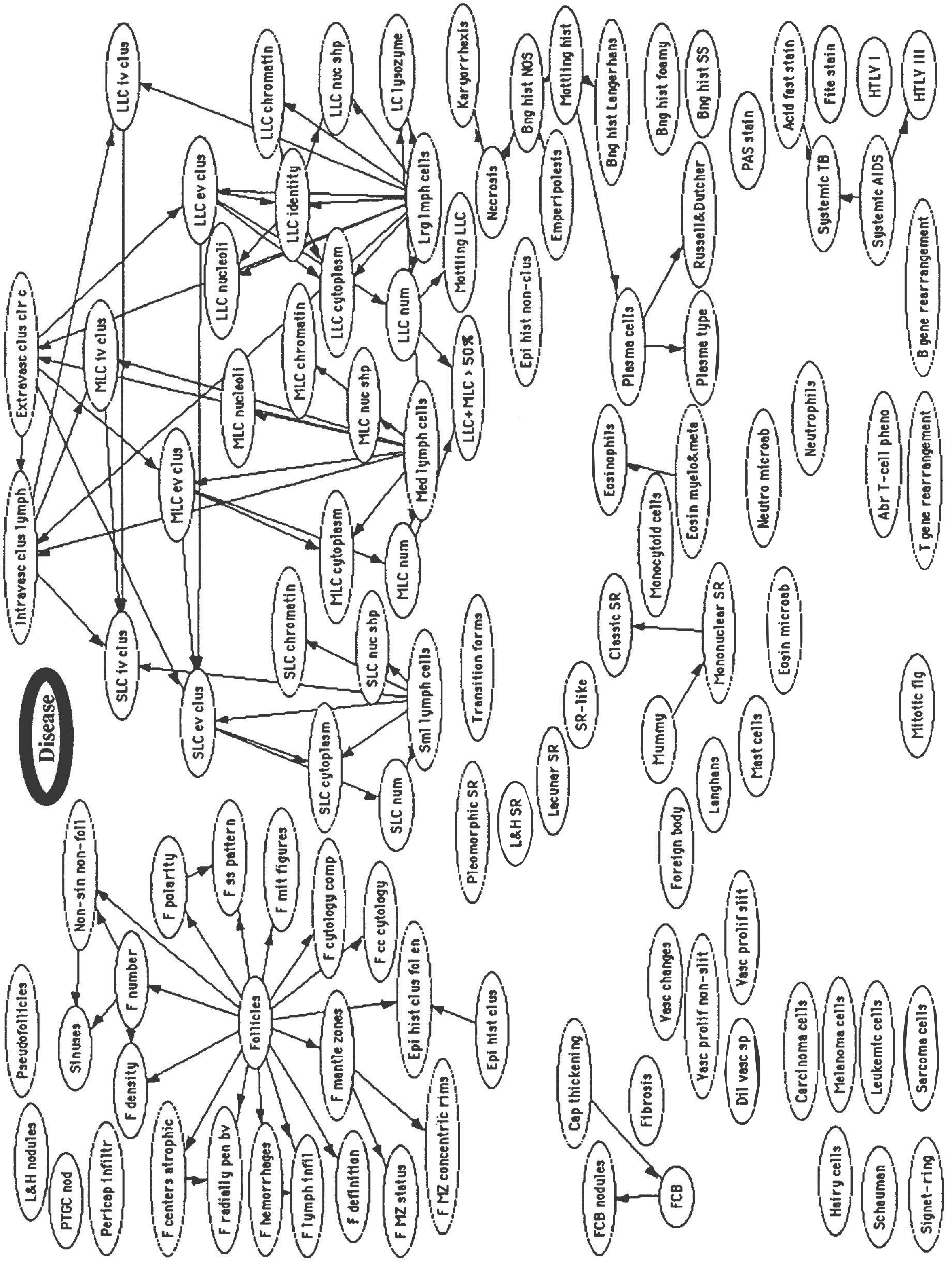

**Figure 1.** The complete belief network for Pathfinder. The node *Disease* contains 63 lymph-node diseases. The arcs from the disease node to all feature nodes are not shown so that the conditional dependencies among features are highlighted. For a complete index to feature and disease abbreviations, see [Heckerman, 1990b].

Although the performance of the CTP method for inference in Pathfinder was satisfactory, we realized that we could improve inference times by taking into account the special structure of the Pathfinder belief network. The disease node in the Pathfinder network is a parent to all but two feature nodes in the network. As a result, the disease node is in almost every clique in the clique tree. Due to the large number (63) of possible values of the disease node, the average state-space size of the cliques is fairly large. By treating the disease node as though it is instantiated, we can remove it from the cliques, simplifying propagation of evidence considerably.

Simplifying network structure by instantiating a node for which no evidence has been observed was first described by Pearl [Pearl, 1986a]. He developed this method, called *cutset conditioning (CC),* to enable the use of Kim and Pearl's polytree algorithm [Kim, 1983; Pearl, 1986b] for inference in multiply connected belief networks. One of the principal drawbacks of Pearl's use of cutset conditioning is that we must cut every loop in the network with an instantiated node (a member of the *loop cutset)* to use the polytree algorithm for inference [Suermondt, (in press)]. During inference, we must consider each possible combination of instantiated values of the loop-cutset nodes; the number of these *loop-cutset instances* is equal to the product of the numbers of possible values of the loop-cutset nodes, and this product is clearly exponential in the number of loop-cutset nodes.

## 2. Aggregation After Decomposition

To use the polytree algorithm for inference in a multiply connected belief network, we must cut every loop in the network. If there are many loops, this requirement leads quickly to a number of loop-cutset instances that is so large that inference becomes intractable. The solution to this problem is to use the CTP method, rather than the polytree algorithm, for inference in the partially decomposed network: After decomposing the loop-cutset nodes, we aggregate the nodes of the revised network into cliques and form a clique tree, which we use for inference. Since CTP does not require that the network be singly connected, we do not need to cut all the loops in the network; rather, we make a loop cutset of one or a few nodes to simplify the structure of the clique tree. The desired result is that the cliques in the revised network are smaller than are those in the original network.

We call this approach *aggregation after decomposition (AD);* we describe it in more detail in [Suermondt, 1990; Heckerman, 1990a, Ch. 4]. In the case of the Pathfinder network, we decompose the disease node by acting as though that node has been instantiated. The loop cutset consists of only the disease node; each instance of the loop cutset corresponds to a single diagnosis, and the weight of each loop-cutset instance [Pearl, 1986a; Pearl, 1988, p. 206; Suermondt, 1989] is equal to the marginal probability of the corresponding diagnosis. After making the disease node the loop cutset, we remove this node from the network and aggregate the remaining nodes to form a clique tree.

Due to the structure of the Pathfinder network, removal of the disease node disconnects sections of the belief network from one another. After we condition on the disease node, the resulting clique tree is not connected. Rather, it consists of several independent portions that can be updated independently. This disconnected clique tree speeds up inference in the case of the Pathfinder network: During inference, we must propagate evidence through only those portions of the clique tree in which there is new evidence. In the original—connected—clique tree, every evidence observation was propagated to the *entire* clique tree. As we shall discuss in Section 3, disconnecting the clique tree is the primary benefit of applying the AD method to the Pathfinder belief network.

During inference using the AD method in Pathfinder, we must consider separately each possible diagnosis. Analogously to cutset conditioning, we maintain a separate copy of the network for each possible diagnosis $d_i$. We maintain a weight $w_i$ that reflects the current marginal probability of diagnosis $d_i$. Initially, $w_i$ is equal to the prior marginal probability of $d_i$, which can be obtained directly from the knowledge base: $w_i = P(d_i)$.

When we observe evidence, we must update the weights to obtain the posterior probability of each diagnosis. Let us denote the new evidence by $E$; $E$ is a set of nodes, each with a single observed value. The new weight for each diagnosis, $w_i'$, is calculated as follows [Pearl, 1987; Pearl, 1988; Suermondt, 1989]:

$$
\begin{aligned}
w_i' &= P(d_i \mid E) \\
&= \alpha P(E \mid d_i) P(d_i) \\
&= \alpha P(E \mid d_i) w_i,
\end{aligned}
$$

where $\alpha$ is the normalization constant over all diagnoses $d_i$; that is, $\alpha = 1/P(E)$.

Clearly, we must know $P(E \mid d_i)$ to be able to calculate $w_i'$. Fortunately, CTP provides us with this probability in a straighforward manner. During inference, we propagate the new evidence through the clique tree for each possible diagnosis. We must propagate evidence through only those portions of the clique tree that contain nodes for which new evidence has been observed. We collect evidence into a single *top clique* for each relevant disconnected portion of the network. Normalizing the marginal probabilities for the top clique of a network portion yields a normalization constant that is equal to the joint probability of all the evidence that is newly observed for that network portion [Lauritzen, 1988, p.185]. By multiplying these normalization constants for all the independent network portions that contain new evidence, we obtain $P(E \mid d_i)$. Thus, we can obtain the posterior marginal probability of each diagnosis given the observed evidence.

### 3. Results of Application of AD to Pathfinder

Removing the disease node from the network results in a clique tree in which the cliques are considerably smaller than are those in the original network. Since the disease node has 63 possible values, and the disease node is a member of virtually all cliques, the average state-space size decreases by a factor of approximately 63 when we remove the disease node. However, this reduction is largely compensated for by the fact that we have to consider separately each of the possible 63 diagnoses. Thus, even though a single evidence-propagation sweep through the network is much less complex, when we observe new evidence, we have to perform 63 of these sweeps (using the AD method), rather than a single one (using CTP). Thus, the net result—the total inference time—may be virtually unchanged: The reduction of the size of the cliques alone does not lead to faster evidence propagation.

Nonetheless, on average AD leads to faster evidence propagation in the Pathfinder network. Due to the special structure of the Pathfinder belief network, instantiation of the disease node yields a disconnected clique tree.[2] In this disconnected clique tree, we need to propagate evidence through only those portions of the clique tree that contain nodes for which new evidence has been observed. When there is evidence in every disconnected portion, the inference time for AD is about the same as for CTP. Usually, however, evidence is present only in some portions of the tree. In these cases, AD is faster than CTP. In the best case, all evidence nodes belong to a single small portion of the clique tree, and all other portions can be left unchanged; in the worst case, each portion of the clique tree contains at least one evidence node, and the entire clique tree must be updated. Notice that, even in the worst case, the AD method is no slower than is CTP.

Clearly, the degree to which inference in Pathfinder is faster when we apply AD rather than CTP depends strongly on the location of the evidence nodes in typical Pathfinder cases. We performed a preliminary evaluation of AD, in which we selected 20 cases in sequence from a large library of referrals. For each case, a community pathologist reported salient morphologic features to Pathfinder. The pathologist entered features until she believed that no additional observations were relevant to the case.

The number of observed features in our test cases averaged 5.7 (range 3–10; $\sigma = 2$). We found that the average run time using AD in the test cases was 0.530 of the run time using CTP in Pathfinder (range 0.098–0.921; $\sigma = 0.389$). Thus, assuming that the cases in our preliminary evaluation represent typical Pathfinder cases, we can expect that inference times using AD will be approximately twice as fast, on average, as those using CTP. However, the large standard deviation $\sigma$ indicates that the improvement in inference time using AD rather than CTP varies widely among the cases.

Strikingly, we found that the run times were clearly clustered (see Figure 2). One cluster contained 9 cases with 3 to 5 features each, and run times that averaged 0.112 of the full CTP run time (range 0.098–0.169; $\sigma = 0.024$); the other cluster had 11 cases with 4 to 10 features each and run times that averaged 0.872 of full CTP run time (range 0.822 - 0.921; $\sigma = 0.032$). Thus, depending on the case, AD performed inference either an order of magnitude faster than did CTP, or only marginally faster.

[2] Strictly speaking, the disconnected clique structure is no longer a tree, but rather a *forest*. In this paper, however, we shall refer to the clique structure—disconnected or not—as the clique tree.

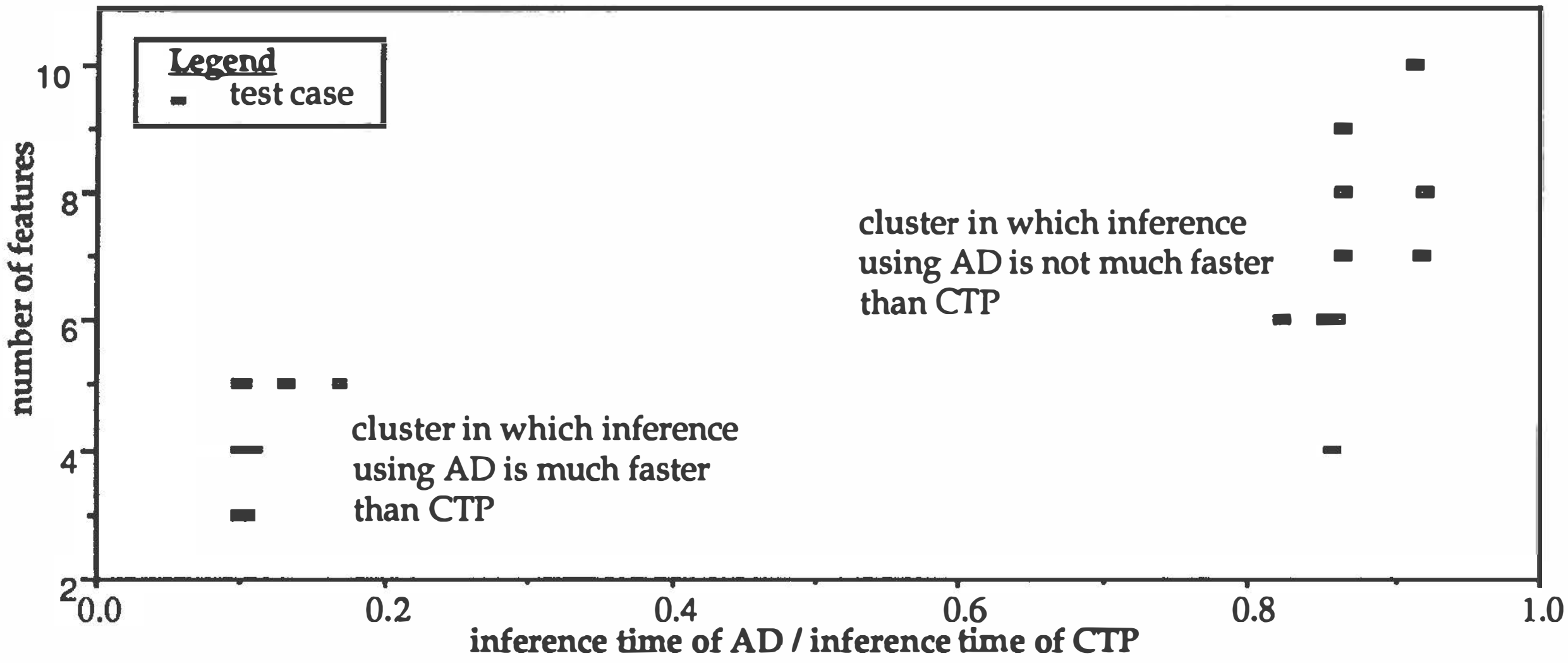

**Figure 2.** Scatterplot of Pathfinder case results. For each test case, we show the number of features along the Y axis and the ratio of inference times using AD and CTP along the X axis. The cases are clearly clustered.

We can explain the clustering of case results by considering the size of the portions into which the clique tree is divided. Many of the independent portions consist of a single node or a pair of nodes; evidence in such portions can be absorbed extremely rapidly. On the other hand, the largest independent portion of the clique tree contains 23 cliques and 29 feature nodes. All cliques containing more than three nodes—including one clique that contains six nodes—are in this network portion. As mentioned in Section 1, the time complexity of evidence propagation between cliques increases exponentially with the number of nodes in each clique. Thus, the inference time required to update the single network portion containing all the largest cliques is greater than that for all the remaining network portions combined. If evidence is observed for any features in that large network portion, inference time using AD will not be much faster than that for CTP. In Section 4, we discuss briefly the possibility of breaking up such large network portions.

## 4. Extensions and Discussion

Selection of an appropriate loop cutset is not a problem in the Pathfinder network: The disease node forms an obvious choice. Clearly, it is possible to consider instantiation of additional nodes as loop-cutset members [Suermondt, 1990; Heckerman, 1990a]. Such an addition might yield a clique tree that is disconnected even further, allowing us to obtain more efficient evidence propagation on average. Although there are several nodes in the Pathfinder network that are suitable loop-cutset candidates, there is an important drawback associated with addition to the loop cutset of nodes other than the disease node. Our primary goal in Pathfinder is to obtain the marginal probability of each diagnosis; if the disease node is the only loop-cutset node, these marginals are equal to the weights of the diagnoses. If we add other nodes to the loop cutset, we must perform additional marginalization operations to obtain the marginal probabilities of the diagnoses after absorption of evidence. Thus, when we add other nodes to the loop cutset, inference using AD may be slower—in the worst case—than inference using CTP.

The method of conditioning, regardless of the manner in which evidence is absorbed, opens the way for several modifications of the inference scheme. One advantage of the conditioning approach is that we can consider each of the diagnoses separately. Thus, evidence propagation for the diagnoses can be performed in parallel, after which we aggregate the results by normalizing the weights. Such a parallel implementation may lead to substantial savings in inference time.

Another option made possible by the use of conditioning in AD is *bounded conditioning* [Horvitz, 1989]. In brief, bounded conditioning allows us to speed up inference by removing from consideration diagnoses with a very low marginal probability. After eliminating these diagnoses, we obtain exact bounds on the posterior probabilities of the remaining diagnoses. The reduction in inference time is related linearly to the number of eliminated diagnoses, provided that we can obtain the joint probability of the new evidence at low computational cost. This limits the applicability of the method to cases where all new evidence is contained in a single clique. A possible danger of applying bounded conditioning is that we remove from consideration the most probable diagnosis. Such a situation could occur if this diagnosis, before instantiation of the evidence nodes, has a very low prior probability. Nonetheless, the wide bounds on the posterior probabilities of the remaining diagnoses will inform us of the situation.

We conclude from our experiments with the Pathfinder network that aggregation after decomposition provides a means by which we can reduce inference times in the Pathfinder network substantially. AD is a hybrid approach between Pearl's conditioning method and Lauritzen and Spiegelhalter's clique-tree propagation. Additional hybrid approaches that combine exact probabilistic-inference algorithms are possible. Such approaches may provide the key to tailoring inference algorithms to the structure of the belief-network knowledge base of an expert system.

## Acknowledgments

We thank Stig K. Andersen for guidance on practical aspects of this research, and for use of the HUGIN® system to obtain a clique tree for the Pathfinder network. We also thank Eric Horvitz and Ross Shachter for interesting discussions about this work. Lyn Dupré and the referees gave helpful comments on earlier versions of this document. Support for this work was provided by the National Science Foundation under grant IRI-8703710, by the U.S. Army Research Office under grant P-25514-EL, and by the National Library of Medicine under grant LM-04529. Computer facilities were provided in part by the SUMEX-AIM resource under grant LM-05208 from the National Institutes of Health.

# *Session 6:*

## Possibility Theory: Semantics and Applications

Possibility as Similarity: the Semantics of Fuzzy Logic
*E. Ruspini*

Integrating Case-Based and Rule-Based Reasoning: the Possibilistic Connection
*S. Dutta and P.P. Bonissone*

Credibility Discounting in the Theory of Approximate Reasoning
*R.R. Yager*

Updating with Belief Functions, Ordinal Conditioning Functions and Possibility Measures
*D. Dubois and H. Prade*